\documentclass{article}
\pdfoutput=1
\usepackage{ijcai19}

\usepackage{times}
\usepackage{soul}
\usepackage{url}
\usepackage[utf8]{inputenc}
\usepackage[small]{caption}
\usepackage{graphicx}
\usepackage{amsmath}
\usepackage{amssymb}
\usepackage{booktabs}
\usepackage{algorithm}
\usepackage{algorithmic}
\usepackage{multirow}
\usepackage{svg}
\usepackage{footnote}
\makesavenoteenv{tabular}
\makesavenoteenv{table}

\title{A Context-Aware Citation Recommendation Model \\ with BERT and Graph Convolutional Networks}

\author{
Chanwoo Jeong$^1$\and
Sion Jang$^1$\and
Hyuna Shin$^1$\and
Eunjeong Park$^2$,\footnote{Co-corresponding Authors} \and
Sungchul Choi$^1$,\footnotemark[1]\\
\affiliations
$^1$ TEAMLAB, Gachon University, Republic of Korea\\
$^2$ Papago, NAVER, Republic of Korea\\
\emails
\{chanwooda0223, wayterren, hyuna1615\}@gmail.com,
lucy.park@navercorp.com,
sc82.choi@gachon.ac.kr
}

\begin{document}

\maketitle

\begin{abstract}

With the tremendous growth in the number of scientific papers being published, searching for references while writing a scientific paper is a time-consuming process. A technique that could add a reference citation at the appropriate place in a sentence will be beneficial. In this perspective, context-aware citation recommendation has been researched upon for around two decades. Many researchers have utilized the text data called the context sentence, which surrounds the citation tag, and the metadata of the target paper to find the appropriate cited research. However, the lack of well-organized benchmarking datasets and no model that can attain high performance has made the research difficult.
 
In this paper, we propose a deep learning based model and well-organized dataset for context-aware paper citation recommendation. Our model comprises a document encoder and a context encoder, which uses Graph Convolutional Networks (GCN) layer and Bidirectional Encoder Representations from Transformers (BERT), which is a pre-trained model of textual data. By modifying the related PeerRead dataset, we propose a new dataset called FullTextPeerRead containing context sentences to cited references and paper metadata. To the best of our knowledge, This dataset is the first well-organized dataset for context-aware paper recommendation. The results indicate that the proposed model with the proposed datasets can attain state-of-the-art performance and achieve a more than 28\% improvement in mean average precision (MAP) and recall@k.

\end{abstract}

\section{Introduction}

Have you ever had difficulty citing references while writing a scientific paper?  Most artificial intelligence researchers may have thought about finding a solution for this problem a little easier at least once. As shown in Figure 1, a possible solution is to automatically find the information that should be cited in a placeholder, such as [REF].

\begin{figure}[h]
    \centering
    \includegraphics[width=\columnwidth]{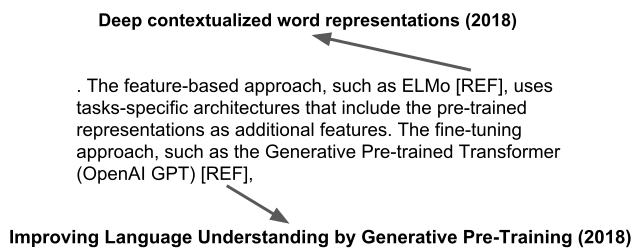}
    \caption[An example manuscript with citation placeholders to find fittable references. This text is from BERT paper]{An example manuscript with citation placeholders to find fittable references. This text is from BERT paper~\cite{2018arXiv181004805D}}
    \label{fig:example}
\end{figure}

Finding a suitable scientific document--considering the surrounding text--that can be used as a placeholder is called "Context-aware citation recommendation"~\cite{he2010context}. The sentences on both sides of a placeholder are referred to as "context". A context-aware citation recommendation task is a type of supervised classification that is used to choose a paper suitable as the placeholder based on its contents. In addition to context, by considering the characteristics of a scientific document, the task has conducted with using author, title, citation, journal (or conference name), etc., which are metadata, or bibliometrics, of a scientific paper~\cite{he2010context,he2011citation,7279056,Fontoura2013ASL,Tang:2014:CCC:2600428.2609564}. In recent years, there has been an increasing number of attempts to solve problems with the use of a deep neural net; this resulted in an increase in the number of articles~\cite{huang2015neural,Ebesu:2017:NCN:3077136.3080730,8478136}.

Even though the task has been a relatively constantly researched area, one of the most challenging aspects of this study is that there is no benchmarking dataset against which proper performance can be measured. In general, this task needs to use metadata along with the context surrounding the cited paper. To the best of our knowledge, suitable datasets have not been disclosed. Among the commonly used data, the ACL Anthology Network (AAN)~\footnote{http://clair.eecs.umich.edu/aan/index.php} dataset does not provide paper sentences and metadata in a preprocessed form, and the DBLP dataset~\footnote{https://dblp.uni-trier.de/xml/} only provides bibliographic information. In a recently published~\cite{huang2015neural}, CiteseerX datasets~\footnote{https://psu.app.box.com/v/refseer} only provided context and citation information and did not provide meta information simultaneously. As a result, related studies have failed to use the same benchmarking dataset.

The purpose of this study is to provide datasets and state-of-the-art models suitable for the context-aware paper recommendation task research, and in turn provide researchers with an improved paper writing environment.
The main contributions of this study are as follows: First, we built reproducible benchmarking datasets for the task. We preprocessed the existing AAN dataset~\cite{aan2013,Radev&al.09a,Radev&al.09.journal} to fit the task, and constructed new dataset called FullTextPeerRead using PeerRead~\footnote{https://github.com/allenai/PeerRead}~\cite{kang2018dataset}. Second, we constructed the state-of-the-art model for the task using BERT~\cite{2018arXiv181004805D} and Graph Convolution Networks (GCN)~\cite{kipf2016semi}. Because scientific papers contain textual contents data, and metadata that can be represented as a graph, we use BERT, which recently proved to have the highest performance level in the field of Natural Language Processing (NLP) for textual data, and GCN for network-based metadata. Finally, we investigated various factors to affect task performance through experiments.

\section{Proposed Dataset}
\subsection{Dataset overview}
We constructed two new datasets for the context-aware citation recommendation task. We suggested revisions of the existing datasets, AAN~\cite{aan2013} and FullTextPeerRead which is an expansion of PeerRead dataset~\cite{kang2018dataset}. AAN and PeerRead datasets have well-organized bibliometric information. The PeerRead dataset mainly provides peer reviews of submitted papers in top-tier venues of the artificial intelligence field, along with bibliometric information. Since all existing datasets of the task lack information in the citation context, our focus is on gathering context information with metadata. Therefore, the AAN and PeerRead datasets were reprocessed to create the dataset.

\subsection{Data acquistion}
We used arXiv Vanity~\footnote{https://www.arxiv-vanity.com/} to create the dataset. arXiv Vanity is a site that converts a latex-based PDF file to HTML document. Our goal is to extract the context information on both sides of the citation symbol, such as [1] or [Choi et al. 2016], together with the reference paper information. To do this, we parsed latex into HTML via arXiv Vanity, and used a regular expression to recognize the citation symbol in the document. Next, the sentences on both sides of the citation symbol were stored on a database with reference article information. We stored the collected information together with the existing metadata and build it into a database.

The data actually collected was noisy because the latex documents were not consistent format. After we automatically collected the necessary data, we removed the noisy data manually. For example, in the case of CiteseerX~\footnote{https://psu.app.box.com/v/refseer}, the citation symbol corresponding to the placeholder is left in the context and data is provided. However, in this case, the placeholder text itself was used for overfitting learning, so the text can be used to tell the correct answer. Exactly, the reminded placeholder can be used as crucial evidence of prediction. Therefore, our final work was post-processed manually because this noise may remain when mechanically collected.

\subsection{Statics of dataset}

Finally, the statics of the dataset we built are as shown in Table~\ref{tab:dataset_statics}. The number of extracted datasets were less than the original number of AAN or PeerRead dataset because we needed to remove paper the PDFs that did not use latex or were very noisy from being processed with arXiv Vanity. In Table ~\ref{tab:dataset_statics} below, Total "\# of base papers" is a paper that cites other researches. We have metadata information of the paper that is used as an input for the classification task. "\# of cited papers" is a cited paper. In addition, we extracted paragraph units on both sides of the citation symbol, and "\# of citation context" means the sum of the number of sentences which are in the extracted paragraph. Further, "\# of total papers" refers to the total number of papers covered by base paper and cited paper, excluding duplicates.

\begin{table}
\centering
\begin{tabular}{ccc}
\hline
Dataset name  & AAN\footnote{http://bit.ly/2IDam8J} & FullTextPeerRead\footnote{http://bit.ly/2Srkdht} \\
\hline
\# of total papers       & 7,073   & 4,898     \\
\# of base papers      & 5,576  & 3,761     \\
\# of cited papers    & 2,417  & 2,478     \\
\# of citation context   & 12,125  & 17,247    \\
\# Paper published year   & 1965 - 2015  & 2007 - 2017     \\
\hline
\end{tabular}
\caption{Dataset description. Because the AAN policy prohibits disclosure of modified data, we cannot disclose it. We will open the dataset after receiving a grant from the AAN}
\label{tab:dataset_statics}
\end{table}

\section{A BERT-GCN Model for Context-Aware Citation Recommendation}

\subsection{Model overview}

We construct the context-aware citation recommendation model using BERT~\cite{2018arXiv181004805D} and GCN~\cite{kipf2016semi}. BERT is one of the highest performing pre-trained models for NLP learning representation. We expect that the learning presentation of context sentences, through pre-trained BERT, will achieve a high performance. Scientific data, such as papers, also contain various metadata, in addition to textual data. We use the GCN model to represent the citation relationship between papers and to extract a learning representation of them. 

\begin{figure}[h]
    \centering
    \includegraphics[width=\columnwidth]{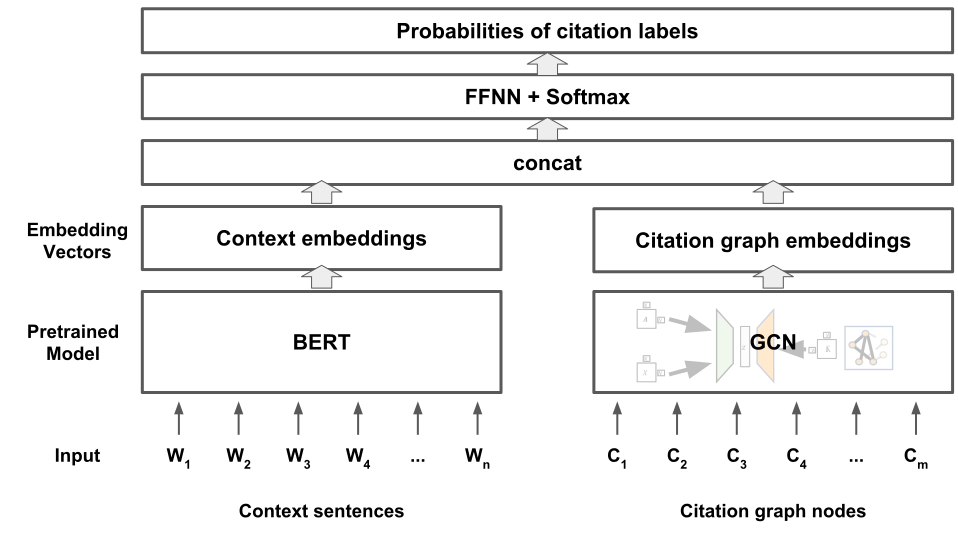}
    \caption{A BERT-GCN model architecture}
    \label{fig:archi}
\end{figure}

As shown in Figure \ref{fig:archi}, we construct a context encoder to extract textual embedding, using BERT, and a citation encoder to extract graph embedding from GCN. Each encoder is pre-trained with context data, and citation graph data is extracted from the paper. Then data is inserted into the pre-trained models, and concatenated embeddings are calculated by each encoder. Finally, after passing the concatenated vectors to a feedforward neural net, the softmax output layer is generated, and cross entropy is adopted as a loss function for training.

The structure of the proposed model is linked to the baseline CACR~\cite{8478136}. CACR has both a paper encoder and a citation context encoder. CACR demonstrates the performance of SOTA as the most recent context-aware citation recommendation model using an AAN dataset and an LSTM model. In the CACR model, a paper encoder was constructed using author, venue, and abstract information in the paper. Our model constructed the citation encoder with GCN solely using citation information.

\subsection{Citation encoder}

The citation encoder conducts unsupervised learning for citation, linking a prediction with the GCN-based Variational Graph Auto-Encoders (VGAE) model~\cite{kipf2016variational}  by using citation relationships between papers as input values. When paper information is used as input to a pre-trained GCN, the model returns the relational learning representation as the embedding vector. VGAE can capture the latent learning representation of graph data.

In existing research, it has been a challenge to convey the citation relationship of a paper as Doc2Vec~\cite{le2014distributed} has been used to encode paper information and summarize it after embedding the learning of the individual meta-information. Our citation encoder complemented this by using citation linking prediction information as a citation prediction feature.

\subsubsection{Graph Convolutional Network (GCN) Layer}

In our model, the role of the GCN layer is to abstract the citation network graph information through a convolutional network. The GCN layer is used as an inference model of VGAE. The formula of the GCN layer for VGAE is shown in Equation \ref{equation:gcn_layer}.

\begin{equation}
GCN(X,A) = \tilde{A} ReLU(\tilde{A}XW_{0})W_{1} \\
\label{equation:gcn_layer}
\end{equation}

The proposed model consists of two GCN layers. The layer uses two matrices as input: the identity matrix $X$ and the adjacency matrix $A$, which is an $N$ by $N$ matrix. $N$ is the number of input paper. Learned from the first GCN layer, layer parameter $W^0$ is used as the weight matrix for the second layer. Each layer uses layer-wise propagation.

$\tilde{A}$ is a normalized adjacency matrix based on the diagonal degree matrix D, as shown in Equation~\ref{equation:tilde}.

\begin{equation}
\tilde{A} = D^{-\frac{1}{2}}AD^{-\frac{1}{2}}
\label{equation:tilde}
\end{equation}

\subsubsection{Variational Graph AutoEncoder (VGAE)}

\begin{figure}[h]
    \centering
    \includegraphics[width=\columnwidth]{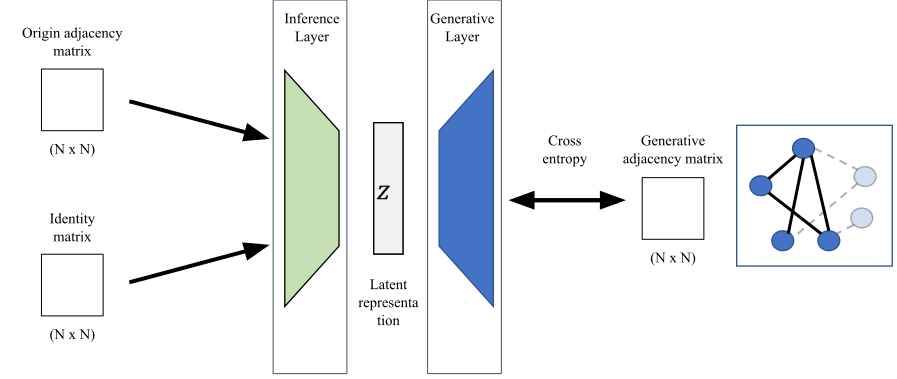}
    \caption{A architecture of Variational Graph AutoEncoder}
    \label{fig:vgae}
\end{figure}

VGAE is a model that applies the unsupervised learning method of the Variational Autoencoder~\cite{Rezende:2014:SBA:3044805.3045035} to the graph structure using GCN. VGAE learns the latent representation by minimizing the cost between inference model and the generative model as shown in Equation~\ref{equation:total_loss}.

\begin{equation}
L = \mathbb{E}_{q(Z|X,A)}[\log p(A|Z)] - KL[q(Z|X,A) || p(Z)]
\label{equation:total_loss}
\end{equation}

The inference layer of VGAE learns the representation Z by to decrease the KL-divergence loss between the normal distribution from the GCN layer result and the Gaussian normal distribution, as shown in Equation~\ref{equation:inference}.

\begin{equation}
\begin{aligned}
q(Z|X,A)    &= \prod\;_{i=1}^{N} q(z_{i} | X, A), q(z_{i} | X, A)  \\
            &= \mathcal{N}(z_{i} | \mu _{i}, diag(\sigma_{i}^{2})) 
\end{aligned}
\label{equation:inference}
\end{equation}

As a next step, the generative layer learns an adjacency matrix based on the representation matrix $Z$ of the interference layer. The latent variable $z_ {i}$, $z_ {j}$ is the inner product value of document $i$ and $j$. An adjacency matrix is generated based on the latent variable through the inner product between paper vectors, as shown in Equation \ref{equation:generative}. The generative model defines the representation matrix $Z$ by reducing the difference between the adjacency matrix $A$ generated by the generative model and the actual adjacency matrix.

\begin{equation}
\begin{aligned}
p(A|Z) &= \prod\;_{i=1}^{N} \prod\;_{j=1}^{N} p(A_{ij} | z_{i}, z_{j}), p(A_{ij} = 1 | z_{i}, z_{j}) \\
    &= \sigma(z_{i}\top z_{j})
\end{aligned}
\label{equation:generative}
\end{equation}

\section{Experiments}
\subsection{Experiments overview}
We compare the proposed model with CACR~\cite{8478136}, one of the existing SOTA models, with a focus on performance. We use the AAN and FullTextPeerRead (FTPR) datasets in our experiments and use Mean Average Precision (MAP), Mean Reciprocal Rank (MRR), and Recall@K as evaluation metrics. The purpose of our experiments is to investigate the following topics, including the overall performance of our model.

\begin{itemize}
\item We compare the performance of the proposed model with the existing SOTA, the CACR model, to gauge the outperformance of BERT and GCN over conventional model. 
\item We investigate differences in performance between models using BERT and GCN. We used BERT for textual data and GCN for graph data. We analyze the impact of each models on overall performance.
\item When using this model in a practical environment to recommend papers, we want to verify that the model can recommend papers solely by looking at the sentence on the left side of a citation symbol. In general, if researchers are writing a paper, researchers expect to recommend reference papers based on the contents which have been written so far without the entire document. We want to check whether our model is available in this situation. 
\item We check the performance of the model according to the length of textual data. When using BERT, we check whether sentences, that are distant from the citation symbol, are used as noise or useful information
\item We measure performance according to paper occurrence in the aggregate dataset. Citation of specific papers is rare; however, we need to understand how our model performs when it happens.

\end{itemize}

\subsection{Experiments setting}

\subsubsection{Experimental dataset}

In the experiment, the AAN dataset used data published before 2014, whereas the FullTextPeerRead dataset comprised paper data published before 2018. After interpreting the data, with sufficient context, and the metadata among datasets, AAN and FullTextPeerRead yielded 6,500 and 4,898 papers, respectively. The datasets were divided into two parts: the AAN dataset used 5,806 pre-2013 papers for the training set, and the remaining 973 others for the test set prior to 2013, and the test set 973 for the test set. Regarding the FullTextPeerRead dataset, 3,411 pre-2017 papers were used for the training set, and 2,559 papers from 2017 were used for the test set. We applied two text lengths, namely 50 and 100, to the citation context features, and measured the related shift in performance. Furthermore, in order to test performance according to the text input characteristics, we conducted an experiment comparing single and pair context characteristics. For this purpose, we used a single context consisting of a 100 text length on the left side of a citation placeholder, and the pair context consisting of a 100 text length on each side of the citation placeholder.

\subsubsection{Evaluation metrics}

For experimental evaluation, we use MAP, MRR, and Recall Top@K, which are general metrics for information retrieval. The MAP measures average precision reflecting the rank position regarding the retrieval list. This indicator is based on the position of the corresponding label values for the K recommendation list, and we measured the indicator with K=30. The MRR indicator is defined as identifying the location of the first occurrence of the actual labels in the recommendation list. Finally, Recall Top@K is defined as an indicator of the actual label hit ratio in a Top@K recommendation list. For this, we evaluate the Recall index through K = {5, 10, 30, 50, 80, 100}

\subsubsection{Parameter setting}
We extracted the embedded context vectors and document vectors from the BERT and GCN layers built in a separate learning process. In BERT, the number of multi-head attention is 12, the encoder stack is 12, the total number of epochs for learning is 30, the batch size is 16, and the Adam optimizer is used. The learning rate is 2e-5, epsilon is 1e-6, beta1 is 0.9, beta2 is 0.999, and the weight decay rate is 0.01. We set the sequence length maximum to 128, padding 0 if the length is shorter than 128, and the hidden size is 768.

With regard to GCN, the number of epochs is 200, the first hidden dimension is same as the document size and the second hidden dimension is 768, the batch size is same as the total document size(full-batch gradient descent), the optimizer is the Adam optimizer~\cite{2014arXiv1412.6980K}, and the learning rate is 0.01

\subsection{Experiments results}
\subsubsection{Baseline Comparison}

As shown in Table~\ref{tbl:base_comparision}, our model delivers a significant performance improvement over the existing CACR. Compared with the SOTA model, all our models showed a three times performance improvement in MAP, MRR, and Recall@K indices, approximately. In particular, Recall@5, which only sees five retrieval citation papers, is a significant improvement. 

In this experiment, both our model and CACR were used solely for papers with a minimum fequency of five citations, and the learning was conducted by considering fifty words on both sides of the citation symbol.

\def\arraystretch{1.5}%
\begin{table*}[t]
\small
    \begin{tabular}{llcccccccc}
        \toprule
            Dataset & Model & MAP & MRR & Recall@5 & Recall@10 & Recall@30 & Recall@50 & Recall@80  \\ \hline        %% column   
            \multirow{5}*{AAN} & BERT-GCN & \textbf{0.6189} & \textbf{0.6036} & \textbf{0.6736} & \textbf{0.7109} & \textbf{0.7814} & \textbf{0.8162} & \textbf{0.8538}  \\ 
            & BERT-GCN-Left & 0.5967 & 0.5818 & 0.6459 & 0.6843 & 0.7506 & 0.785 & 0.8245  \\
            & BERT & 0.6118 & 0.5971 & 0.6593 & 0.6976 & 0.7645 & 0.81 & 0.8257 \\
            & BERT-Left & 0.5928 & 0.5789 & 0.6364 & 0.6678 & 0.7379 & 0.7793 & 0.8203  \\
            & CACR~\cite{8478136} & 0.2893 & 0.2917 & 0.3861 & 0.4531 & 0.5799 & 0.6573 & 0.721 \\ \hline
            
            \multirow{5}*{FullTextPeerRead} & BERT-GCN & \textbf{0.4181} & \textbf{0.4179} & \textbf{0.4864} & \textbf{0.5291} & \textbf{0.6036} & \textbf{0.6495} & \textbf{0.6994}  \\ 
            & BERT-GCN-Left & 0.3883 & 0.388 & 0.4455 & 0.4815 & 0.5539 & 0.5991 & 0.6499 \\
            & BERT & 0.4152 & 0.415 & 0.4801 & 0.52 & 0.5926 & 0.6366 & 0.6887  \\
            & BERT-Left & 0.3823 & 0.3821 & 0.4391 & 0.4755 & 0.5459 & 0.5885 & 0.6392  \\
            & CACR~\cite{8478136} & 0.1551 & 0.1549 & 0.2154 & 0.2761 & 0.4128 & 0.4794 & 0.5516 \\ \hline \bottomrule
    \end{tabular}
    \caption {MAP, MRR, and Recall@K scores for a frequency of over five citations and fifty pair context sentences}
    \label{tbl:base_comparision}
\end{table*}

We compared the performance by independently reproducing the Python code related in the CACR paper. There was no detailed experimental information such as frequency in the actual paper. Since no frequency was mentioned, we assume that the performance described in the CACR paper is based on a frequency of one, and we compare our performance with the performance described in the paper of CACR as shown in Table~\ref{tbl:base_comparision}. For MAP, MRR, and Recal@10, our model outperform, but after Recall@10, it does underperform. Based on experience, we guessed this to be a phenomenon that occurs when the classification label value is returned with a high frequency of cited papers.

\def\arraystretch{1.5}%
\begin{table*}[h]
\small
    \begin{tabular}{llcccccccc}
                \toprule
            Dataset & Model & MAP & MRR & Recall@5 & Recall@10 & Recall@30 & Recall@50 & Recall@80  \\ \hline        %% column   
            \multirow{2}*{AAN} & BERT-GCN & \textbf{0.4516} & \textbf{0.4323} & \textbf{0.4948} & \textbf{0.532} & 0.5839 & 0.6118 & 0.6335 \\ 
            & CACR~\cite{8478136} & 0.301 & 0.335 & - & 0.474 & \textbf{0.615} & \textbf{0.653} & \textbf{0.712} \\ \hline

    \end{tabular}
    \caption {MAP, MRR, Recall@K scores with a frequency of over one citation, 200 pair context sentences and CACR metric result }
    \label{tbl:cacr}
\end{table*}

\subsubsection{Impact of BERT and GCN}

\begin{figure}[h]
    \centering
    \includegraphics[width=\columnwidth]{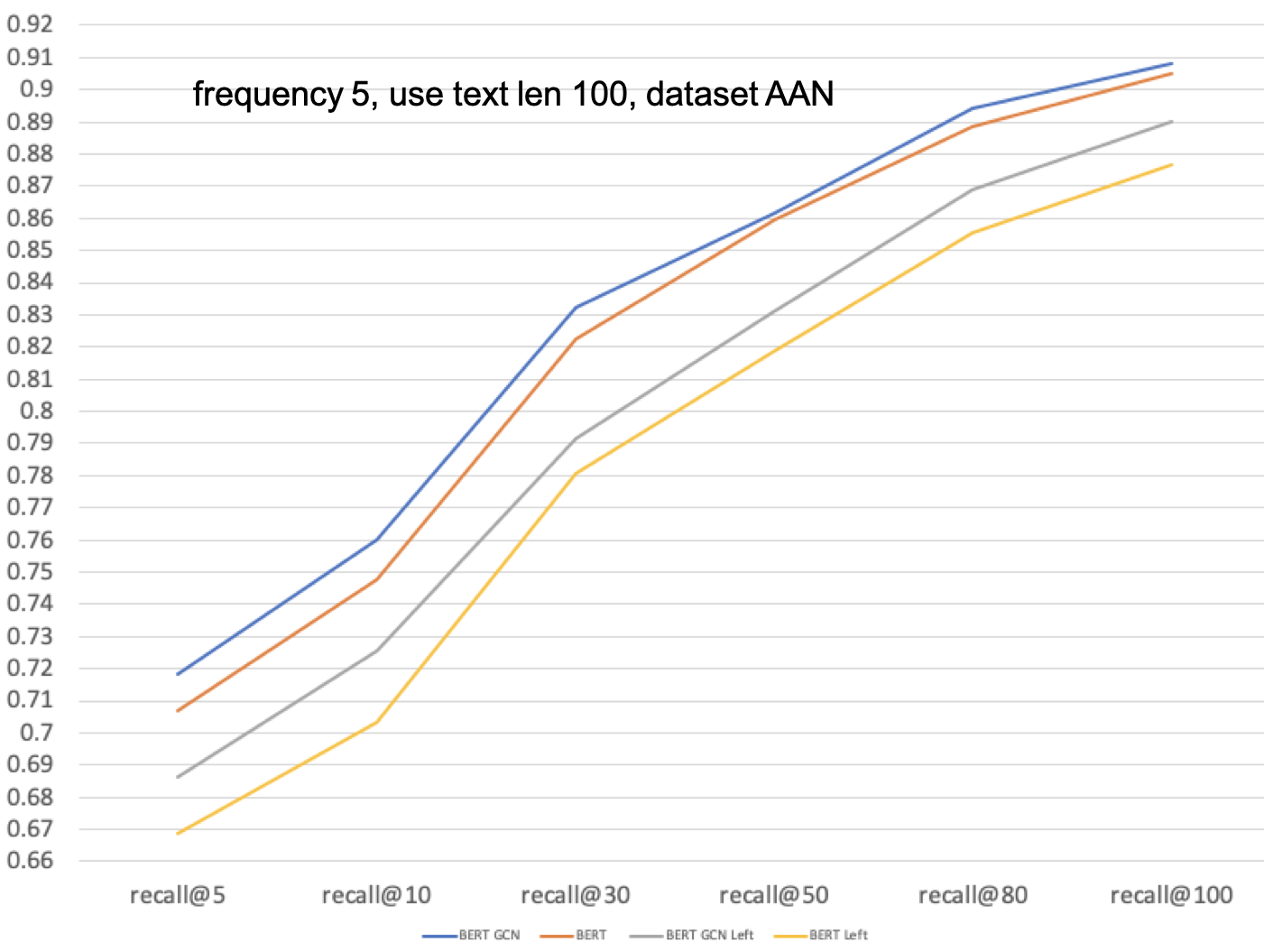}
    \caption{The effects of BERT, GCN and BERT-left}
    \label{fig:bert-gcn}
\end{figure}

In all cases, when the GCN was added, the performance of the model improved. Of interest is the impact that GCN has on the performance of BERT, and BERT-left alone. As shown in Figure~\ref{fig:bert-gcn}, the difference in recall performance between BERT + GCN-Left and BERT-left differs from the difference in performance difference between BERT + GCN and BERT. BERT-left has half the input of context sentence of BERT. Hence, the impact of GCN is greater when the value of the input words is small.

\subsubsection{Impact of the left side context alone }

From the viewpoint of an actual researcher writing a paper, we check the performance when using the context sentence on the left side of the citation symbol alone. As shown in Table\ref{tbl:base_comparision}, the model using the left context alone has a performance that is approximately 0.03 lower than the model using the entire context for the context encoder. In other words, it is helpful to consider the entire context on both sides of the quotation mark. However, adding GCN to the BERT model considering the left context alone delivers performance improvements over adding GCN to the BERT model that considers both contexts, as the network information of the GCN can be helpful when textual information is not sufficient.

\def\arraystretch{1.5}%
\begin{table*}[t]
\small
    \begin{tabular}{|p{3.5cm}|l|p{12cm}|}
        \hline    
        Citation Context & Approaches & Top-3 System \\ \hline         
        \multirow{9}{*}{\shortstack[l]{... inbold means significant \\ ly better than the baseline \\ according to
        \textcolor{red}{\textbf{[?]}} or p-value \\ less than 0.01.baseline SMT \\ system. The decoding ...
}} & \multirow{3}*{BERT-GCN} & 1. Statistical Significance Tests For Machine Translation Evaluation (O) \\      
        & & 2. Bleu: A Method For Automatic Evaluation Of Machine Translation (X)  \\ 
        & & 3. On Coreference Resolution Performance Metrics (X) \\ \cline{2-3}
        & \multirow{3}*{BERT} & 1. Minimum Error Rate Training In Statistical Machine Translation (X)  \\ 
        & & 2. Statistical Phrase-Based Translation (X) \\ 
        & & 3. Statistical Significance Tests For Machine Translation Evaluation (O) \\ \cline{2-3}
        & \multirow{3}*{CACR} & 1. TnT - A Statistical Part-Of-Speech Tagger (X) \\ 
        & & 2. Stochastic Inversion Transduction Grammars And Bilingual Parsing Of Parallel Corpora (X) \\ 
        & & 3. A Maximum Entropy Model For Part-Of-Speech Tagging (X) \\ \hline
        \multirow{6}*{\shortstack[l]{Many researchers have atte \\ mpted to make use of cue p \\ hrases especially for segme \\ ntation both in prose
        \textcolor{red}{\textbf{[?]}}  
}} &  \multirow{3}*{BERT-GCN-Left} & 1. TextTiling: Segmenting Text Into Multi-Paragraph Subtopic Passages (O) \\      
        & & 2. Multi-Paragraph Segmentation Of Expository Text (X)  \\ 
        & & 3. Discourse Segmentation Of Multi-Party Conversation (X) \\ \cline{2-3}
        & \multirow{3}*{BERT-Left} & 1. Multi-Paragraph Segmentation Of Expository Text (X)  \\ 
        & & 2. Advances In Domain Independent Linear Text Segmentation (X) \\ 
        & & 3. TextTiling: Segmenting Text Into Multi-Paragraph Subtopic Passages (O) \\ \hline
    \end{tabular}
    \captionof{table}{A comparison of ground truth with the top five recommended citation lists}
    \label{tab:examples}
\end{table*}

\subsubsection{Effect of length of the context sequence}

\begin{figure}[h]
    \centering
    \includegraphics[width=\columnwidth]{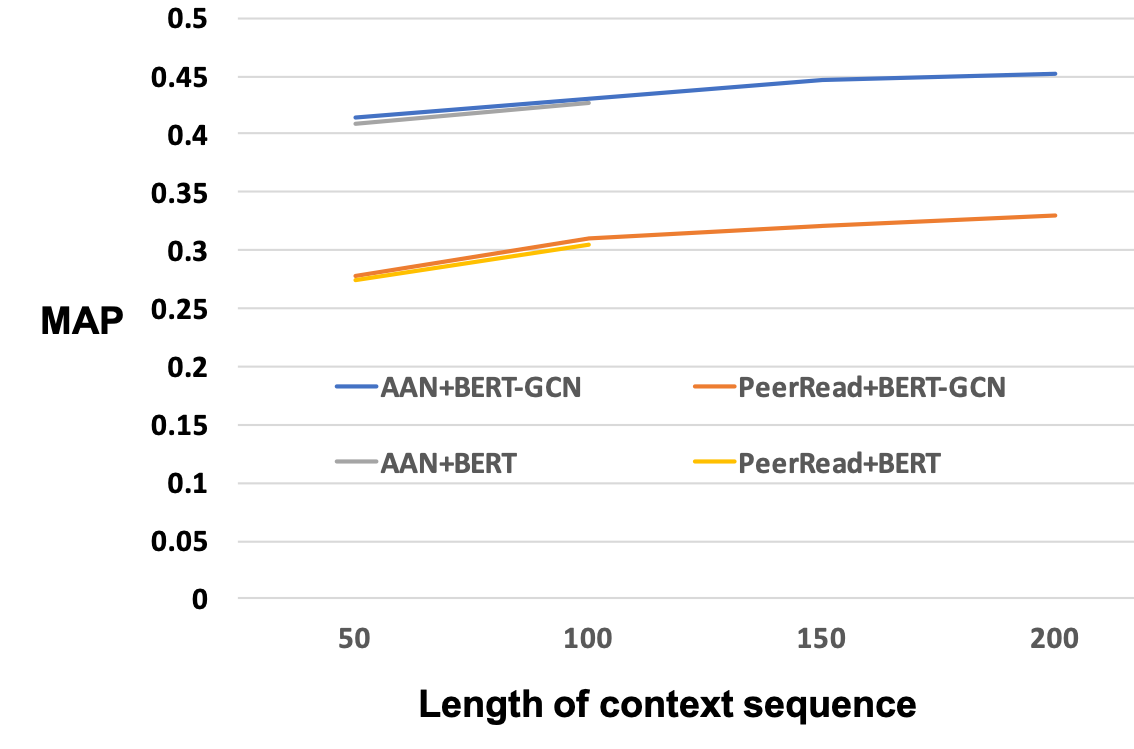}
    \caption{The shift in performance according to the length of sentence context when frequency is one}
    \label{fig:sequence_length}
    \setlength{\belowcaptionskip}{-10pt}
\end{figure}

We have already confirmed a shift in performance tied to the length of the input context sentences. However, how much more can performance improve as the number of sentences increases? As shown in Figure~\ref{fig:sequence_length}, when the context length becomes a hundred or more, model performance becomes less impacted by comparative context length. The context sentence length is relevant, but beyond a certain point, its impact is actually reduced.

\subsubsection{Effect of paper citation frequency}
Finally, we examine the impact of citation frequency on performance. The results of the experiment with citation frequencies one, three, and five show that performance improves when citation frequency is higher, as shown in Table~\ref{tab:freq_comparision}. In general, papers not cited are not used for learning and can be processed as sparse data even at the time of testing. We believe that the learning data should be refined according to citation frequency for a well-functioning, high-performance model.

\begin{table}
\centering
\begin{tabular}{cccc}
\hline
Model  & Frequency 1 & Frequency 3 & Frequency 5 \\
\hline
 BERT GCN  & 0.4467    & 0.6063   & 0.6736     \\
 BERT  & 0.4423 & 0.5971  & 0.6593     \\
\hline
\end{tabular}
\caption{A comparison of shifts in performance based on cited paper frequency}
\label{tab:freq_comparision}
\end{table}

\subsection{Recommendation Examples}

Actual recommendation examples using our model can be found in Table~\ref{tab:examples}. In these examples, we observe that one of the biggest effects of GCN is that it is possible to get a higher relevance similarity of cited paper through GCN for similar articles when we look at textual data alone. Expanding the use of GCN, would not only target similarities in the content of the papers but also citation information in the previous thesis or citation information in the current thesis. With regard to the use of GCN, the representation learning information concerning citation is generated by identifying not only a similarity in the content of the citing paper but also through the citation information of the previous paper by the author or the information in the cited paper of the current article. We think this information is very helpful to improve the performance of the context-aware citation recommendation process.

\section{Conclusion}
The proposed model for context-aware citation recommendation task delivers a significant improvement in MAP, MRR and Recall@K over the existing model. The basis for the breakthrough performance improvement is that the BERT model, which has performed well in recent NLP tasks, is adapted to our context-aware framework. Through the context encoding via BERT, our framework improves the representation learning of the context side. In addition, we apply VGAE, which comprises a GCN layer according to graph data to mitigate over-fitting to local contexts when BERT is applied alone. The VGAE applied to our framework citation encoder processes the paper citation network graph data into a paper latent representation. The combination of the encoded paper network and the encoded context is regularized, resulting in a performance increase over a BERT-based model.

Regarding the context-aware citation recommendation study, existing datasets are not up-to-data and there is no clear context detection. To address this problem, we devised and released the FullTextPeerRead dataset. The proposed dataset comprises updated papers to an extent including papers up-to 2017, and provides a method to readily and accurately extract context metadata and has a well-organized perspective.

\bibliographystyle{named}
\bibliography{main}

\begin{thebibliography}{}

\bibitem[\protect\citeauthoryear{{Devlin} \bgroup \em et al.\egroup
  }{2018}]{2018arXiv181004805D}
J.~{Devlin}, M.-W. {Chang}, K.~{Lee}, and K.~{Toutanova}.
\newblock {BERT: Pre-training of Deep Bidirectional Transformers for Language
  Understanding}.
\newblock {\em arXiv e-prints}, October 2018.

\bibitem[\protect\citeauthoryear{Dragomir R.~Radev}{2009}]{Radev&al.09.journal}
Bryan Gibson Pradeep~Muthukrishnan Dragomir R.~Radev, Mark Thomas~Joseph.
\newblock {A} {B}ibliometric and {N}etwork {A}nalysis of the field of
  {C}omputational {L}inguistics.
\newblock {\em Journal of the American Society for Information Science and
  Technology}, 2009.

\bibitem[\protect\citeauthoryear{Ebesu and
  Fang}{2017}]{Ebesu:2017:NCN:3077136.3080730}
Travis Ebesu and Yi~Fang.
\newblock Neural citation network for context-aware citation recommendation.
\newblock In {\em Proceedings of the 40th International ACM SIGIR Conference on
  Research and Development in Information Retrieval}, SIGIR '17, pages
  1093--1096, New York, NY, USA, 2017. ACM.

\bibitem[\protect\citeauthoryear{Fontoura \bgroup \em et al.\egroup
  }{2013}]{Fontoura2013ASL}
Marcus Fontoura, H.~Asthana, Pradeep~B. Teregowda, Eric Treece, Madian Khabsa,
  Douglas Jordan, Stephen Carman, Prasenjit Mitra, C.~Lee Giles, Bibek Paudel,
  Avishek Anand, Klaus Berberich, St{\'e}phane Marchand-Maillet, Fidel Cacheda,
  Vreixo Formoso, and Victor Carneiro.
\newblock A supervised learning method for context-aware citation
  recommendation in a large corpus.
\newblock 2013.

\bibitem[\protect\citeauthoryear{He \bgroup \em et al.\egroup
  }{2010}]{he2010context}
Qi~He, Jian Pei, Daniel Kifer, Prasenjit Mitra, and Lee Giles.
\newblock Context-aware citation recommendation.
\newblock In {\em Proceedings of the 19th international conference on World
  wide web}, pages 421--430. ACM, 2010.

\bibitem[\protect\citeauthoryear{He \bgroup \em et al.\egroup
  }{2011}]{he2011citation}
Qi~He, Daniel Kifer, Jian Pei, Prasenjit Mitra, and C~Lee Giles.
\newblock Citation recommendation without author supervision.
\newblock In {\em Proceedings of the fourth ACM international conference on Web
  search and data mining}, pages 755--764. ACM, 2011.

\bibitem[\protect\citeauthoryear{Huang \bgroup \em et al.\egroup
  }{2015}]{huang2015neural}
Wenyi Huang, Zhaohui Wu, Chen Liang, Prasenjit Mitra, and C~Lee Giles.
\newblock A neural probabilistic model for context based citation
  recommendation.
\newblock In {\em Proceedings of the Twenty-Ninth AAAI Conference on Artificial
  Intelligence}, pages 2404--2410. AAAI Press, 2015.

\bibitem[\protect\citeauthoryear{Kang \bgroup \em et al.\egroup
  }{2018}]{kang2018dataset}
Dongyeop Kang, Waleed Ammar, Bhavana Dalvi, Madeleine van Zuylen, Sebastian
  Kohlmeier, Eduard Hovy, and Roy Schwartz.
\newblock A dataset of peer reviews (peerread): Collection, insights and nlp
  applications.
\newblock In {\em Proceedings of the 2018 Conference of the North American
  Chapter of the Association for Computational Linguistics: Human Language
  Technologies, Volume 1 (Long Papers)}, volume~1, pages 1647--1661, 2018.

\bibitem[\protect\citeauthoryear{{Kingma} and {Ba}}{2014}]{2014arXiv1412.6980K}
D.~P. {Kingma} and J.~{Ba}.
\newblock {Adam: A Method for Stochastic Optimization}.
\newblock {\em arXiv e-prints}, December 2014.

\bibitem[\protect\citeauthoryear{Kipf and Welling}{2016a}]{kipf2016semi}
Thomas~N Kipf and Max Welling.
\newblock Semi-supervised classification with graph convolutional networks.
\newblock {\em arXiv preprint arXiv:1609.02907}, 2016.

\bibitem[\protect\citeauthoryear{Kipf and Welling}{2016b}]{kipf2016variational}
Thomas~N Kipf and Max Welling.
\newblock Variational graph auto-encoders.
\newblock {\em NIPS Workshop on Bayesian Deep Learning}, 2016.

\bibitem[\protect\citeauthoryear{Le and Mikolov}{2014}]{le2014distributed}
Quoc Le and Tomas Mikolov.
\newblock Distributed representations of sentences and documents.
\newblock {\em arXiv preprint arXiv:1405.4053}, 2014.

\bibitem[\protect\citeauthoryear{{Liu} \bgroup \em et al.\egroup
  }{2015}]{7279056}
H.~{Liu}, X.~{Kong}, X.~{Bai}, W.~{Wang}, T.~M. {Bekele}, and F.~{Xia}.
\newblock Context-based collaborative filtering for citation recommendation.
\newblock {\em IEEE Access}, 3:1695--1703, 2015.

\bibitem[\protect\citeauthoryear{Radev \bgroup \em et al.\egroup
  }{2009}]{Radev&al.09a}
Dragomir~R. Radev, Pradeep Muthukrishnan, and Vahed Qazvinian.
\newblock The {ACL} anthology network corpus.
\newblock In {\em Proceedings, ACL Workshop on Natural Language Processing and
  Information Retrieval for Digital Libraries}, Singapore, 2009.

\bibitem[\protect\citeauthoryear{Radev \bgroup \em et al.\egroup
  }{2013}]{aan2013}
DragomirR. Radev, Pradeep Muthukrishnan, Vahed Qazvinian, and Amjad Abu-Jbara.
\newblock The acl anthology network corpus.
\newblock {\em Language Resources and Evaluation}, pages 1--26, 2013.

\bibitem[\protect\citeauthoryear{Rezende \bgroup \em et al.\egroup
  }{2014}]{Rezende:2014:SBA:3044805.3045035}
Danilo~Jimenez Rezende, Shakir Mohamed, and Daan Wierstra.
\newblock Stochastic backpropagation and approximate inference in deep
  generative models.
\newblock In {\em Proceedings of the 31st International Conference on
  International Conference on Machine Learning - Volume 32}, ICML'14, pages
  II--1278--II--1286. JMLR.org, 2014.

\bibitem[\protect\citeauthoryear{Tang \bgroup \em et al.\egroup
  }{2014}]{Tang:2014:CCC:2600428.2609564}
Xuewei Tang, Xiaojun Wan, and Xun Zhang.
\newblock Cross-language context-aware citation recommendation in scientific
  articles.
\newblock In {\em Proceedings of the 37th International ACM SIGIR Conference on
  Research \&\#38; Development in Information Retrieval}, SIGIR '14, pages
  817--826, New York, NY, USA, 2014. ACM.

\bibitem[\protect\citeauthoryear{{Yang} \bgroup \em et al.\egroup
  }{2018}]{8478136}
L.~{Yang}, Y.~{Zheng}, X.~{Cai}, H.~{Dai}, D.~{Mu}, L.~{Guo}, and T.~{Dai}.
\newblock A lstm based model for personalized context-aware citation
  recommendation.
\newblock {\em IEEE Access}, 6:59618--59627, 2018.

\end{thebibliography}

% \section{\LaTeX{} and Word Style Files}\label{stylefiles}

% The \LaTeX{} and Word style files are available on the IJCAI--19
% website, \url{http://www.ijcai19.org}.
% These style files implement the formatting instructions in this
% document.

% The \LaTeX{} files are {\tt ijcai19.sty} and {\tt ijcai19.tex}, and
% the Bib\TeX{} files are {\tt named.bst} and {\tt ijcai19.bib}. The
% \LaTeX{} style file is for version 2e of \LaTeX{}, and the Bib\TeX{}
% style file is for version 0.99c of Bib\TeX{} ({\em not} version
% 0.98i). The {\tt ijcai19.sty} style differs from the {\tt
% ijcai18.sty} file used for IJCAI--18.

% The Microsoft Word style file consists of a single file, {\tt
% ijcai19.doc}. This template differs from the one used for
% IJCAI--18.

% These Microsoft Word and \LaTeX{} files contain the source of the
% present document and may serve as a formatting sample.  

\end{document}